\documentclass[letterpaper, 10 pt, conference]{ieeeconf}  

\IEEEoverridecommandlockouts                              

\usepackage[pdftex]{graphicx}
\graphicspath{{images/}}
\DeclareGraphicsExtensions{.pdf,.jpeg,.jpg,.png}
\usepackage{mathptmx} 
\usepackage{times} 
\usepackage{amsmath} 
\usepackage{amssymb}  
\usepackage{hyperref}
\usepackage{cleveref}
\usepackage{hyperref}
\title{\LARGE \bf Deployment of Aerial Robots after a major fire of an industrial hall with hazardous substances, a report}

\author{Hartmut Surmann$^{1}$, Dominik Slomma$^{1,3}$, Stefan Grobelny$^{2}$, Robert Grafe$^{3}$
  \thanks{*This work was founded by the Federal Ministry of Education and Research (BMBF) under grant number 13N14860 (A-DRZ https://rettungsrobotik.de/).}
\thanks{$^{1}$University of Applied Science, Gelsenkirchen,  $^{2}$Fire Department of Dortmund, Institute of Fire Service and Rescue Technology, $^{3}$German Rescue Robotic Centre, Dortmund.}
\thanks{corresponding author: {\tt\small hartmut.surmann@w-hs.de}}
\thanks{Further material at: \url{https://github.com/RoblabWh/Berlin}}}

\begin{document}

\maketitle
\thispagestyle{empty}
\pagestyle{empty}

\begin{abstract}
  This technical report is about the mission and the
experience gained during the reconnaissance of an 
industrial hall with hazardous substances after a major fire in Berlin. During this 
operation, only UAVs and cameras were used to obtain
information about the site and the building. First, a
geo-referenced 3D model of the building was created in
order to plan the entry into the hall. Subsequently, the UAVs were used to fly in the heavily damaged interior and take pictures from inside of the hall. A 360° camera mounted under the UAV was used to collect images of the surrounding area especially from sections that were difficult to fly into. Since
the collected data set contained similar images as well as
blurred images, it was cleaned from non-optimal images
using visual SLAM, bundle adjustment and blur detection so that a 3D model and
overviews could be calculated. It was shown that the
emergency services were not able to extract the necessary
information from the 3D model. Therefore, an interactive
panorama viewer with links to other 360° images was implemented where the links to the other images depends on the semi dense point cloud and located camera positions of the visual SLAM algorithm so that the emergency forces could view the surroundings.
\end{abstract}

\section{Introduction}
\label{introduction}

Young scientists in the field of robotics are often asked what their robot, algorithm or sensor is good for and the answer is often for use in the field of rescue robotics. On the one hand, the question in the field of research is unfair, on the other hand, it also shows how little known requirements are in the real use of robots in dangerous situations and that is exactly what we want to improve with this article. In the evening on Thursday, 11.02.2021 a fire in a metalworking factory in Berlin, Germany, could only be extinguished after more than 12 hours of work \cite{berlinfeuerwehr}.
Hazardous substances were released during use.
Due to the stored hazardous substances residents had to keep their windows closed.
A radius of about three kilometers around the fire was affected.
According to the fire brigade, 2,000 square meters of the approximately 5,000 square meter burned initially.
Despite the extinguishing work, the fire then spread to the entire production hall.
Up to 170 firefighters were on duty at the same time.

Due to the high level of damage, the hall was closed and an entry ban was issued.
So, how to get images and information from the inside?


\section {The Job}
\label{sec:mission}

On February 22nd, 2021, the Berlin police submitted an administrative assistance request to the Dortmund fire department for support with special UAV technology as part of a fire investigation.

As a result, a team of emergency services from the Dortmund fire department (FwDO) and staff members from the German Rescue Robotics Centre (DRZ) and Westphalian University of Applied Science were put together and set off to Berlin for a three-day mission with the robotic command vehicle (RobLW).
The order for the team was the digital representation of the outside and inside area of the industrial hall.
In addition to the UAV and image technology, the RobLW was also used on site, with the built-in technology of which the situation images of the UAVs were displayed in real time in order to be able to create corresponding 3D views.
Due to the remaining 10-30 cm high and potentially contaminated extinguishing water on the ground, it was clear from the outset that ground robots could not be used.
The recordings of the image and video material were made available to the Berlin police for assessing the situation and for further investigations.

The use of robotic systems in this dangerous scenario for emergency services was necessary because the dangers for them could not be conclusively assessed from the outside.
By supporting with UAV and special image technology, it was possible for the first time to see the inside of the hall, because of flying the inside of the hall with previous UAV technology by trained UAV pilots was too risky. In addition to the provided 3D views, numerous findings for the further development of robot technology were obtained.

\section {State of the art}
\label{sec:sta}
Rescue robotics encompasses a broad field in research. Because the robots are tested in optimal environments, they rarely find application in real-world operations. Researching the results in real-world conditions is fundamentally important to optimize the robots for specific scenarios. For example, ground robots could not be optimally deployed in a collapsed building in Cologne Germany in 2009, because the debris did not allow the robots to advance \cite{5981550}. Two years later, in 2011, there were major earthquakes in Japan, whereupon ground robots were used to investigate hazardous areas \cite{japan}. In addition to ground robots, underwater robots were also used to find missing people and explore other areas. In 2012 and 2016, there were strong earthquakes in Italy, after which buildings were in danger of collapsing \cite{Kruijff-amatrice, 6523866, advanced-robotics-2014}.There, ground and aerial robots were used to investigate building damage. Since these robots also could not penetrate all areas, water robots were used to identify the remaining areas. These missions show the advantages of using robots and reduce the risk of humans having to put themselves in danger. However, it is clear from the missions that the connection of the research to the real missions is dependent on the force used and how it is trained. Therefore, different real-world environments are used to optimize the robots and train the responders \cite{7017681}. For example, the Fukushima site provides an optimal test site for testing the use of robots \cite{9419563}. In addition to the missions and test areas, it is noticeable that the use of UAVs is fundamentally important in the rescue sector. UAVs make it possible to observe a large area in a very short time and quickly obtain an overall view \cite{mayer:hal-02128385, surmann2019ssrr}.


\section{Deployment}

\subsection{Hardware and software equipment}

\subsubsection{Robotic Command Vehicle (RobLW)}
The German Rescue Robotics Centre
(DRZ) and the Dortmund Fire Department (FwDO) cooperate in using the RobLW. The DRZ uses the vehicle primarily for research, to analyse in which areas of operations the robots can support the emergency forces in a meaningful way. By supporting aerial and ground robots, it is possible to enter and investigate areas without putting people in danger. In addition
to research, the RobLW is used by the FwDO to test the introduced findings
in real-life scenarios. For this purpose, the DRZ works closely with the Dortmund fire department.
The RobLW is 7 m long, 2 m wide and 3 m high and can be divided into three areas. 
\begin{figure}[!t]
\centering
\includegraphics[width=0.49\textwidth]{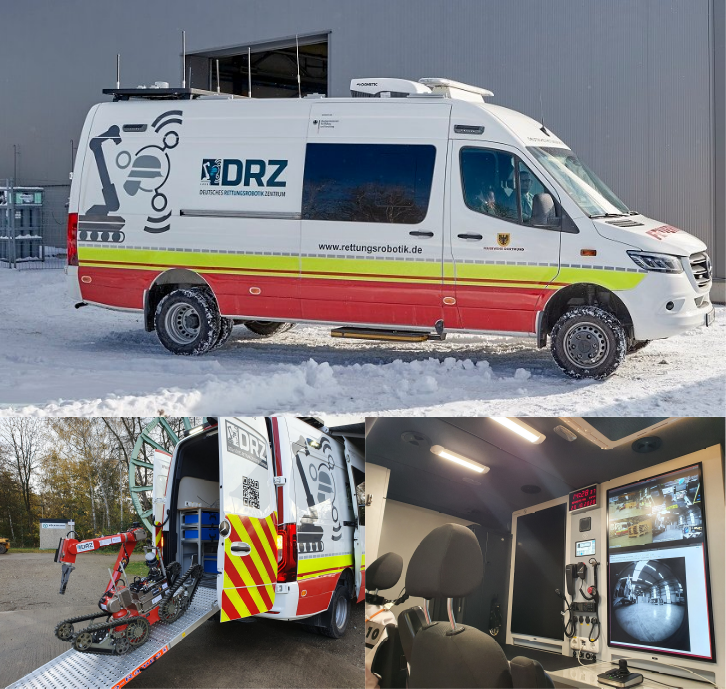}
\caption{Overview of the RobLW. Top: The RobLW. Bottom left: Trunk with a ground robot. Bottom right: The command center with two monitor workstations.}
\label{fig:rlw-overview}
\end{figure}
The front area is the command compartment which is used to control the vehicle. In addition, it is possible to contact the command centre and other emergency units by radio from there. It is also possible to connect to the vehicle's network with a separate computer to share information or access the vehicle's infrastructure.  In the rear area, various peripheral equipment and components are stored for use. Furthermore, the compartment provides a storage area for ground robots and UAVs to be transported.  The middle area represents the operation centre. This is where the robots and UAVs are controlled and where the firefighters can be supported by suitable assistance systems. For this purpose, the operation centre of the RobLW is equipped with hardware for two seats and a powerful server. A screen with an Intel NUC and the associated peripherals is available for each operator. The Intel NUCs are used exclusively for the visualization of camera data or the visual representation of, for example, 3D created environments. The calculations for the creation of three-dimensional environments or artificial neural networks take place on the server. This has an AMD Ryzen Threadripper processor with 64 cores, 64 GB RAM and an Nvidia Geforce RTX 2080 with 8 GB dedicated memory. To ensure communication between the server and the NUCs, a special TDT router is used. This not only takes care of the communication, but also enables access to the Internet via the mobile network. It offers the possibility to download local maps of the environment during an operation. In addition, current data can be sent to the operations control centre to get a better overview of the current situation. Furthermore, the router creates its own Wi-Fi network. This offers the option that the operation commander does not have to evaluate the data directly in the RobLW but can be closer to the operation site if it is connected to the RobLW. The operations centre of the RobLW also has its own radios, for example, to coordinate with other UAV pilots or to provide the operations commander with current details as quickly as possible.

Since the server is mainly responsible for processing the various data from the robots, it contains the primary software to support the fire department. To ensure that the operators know where the robots are at all times, the server contains a so-called situation image system. This receives data from the robots, such as the camera data and the GPS information of the individual robots. This information is used to track the robots in a client software in real time and to determine their position. Furthermore, the information of each robot can be viewed and analysed. In addition to the positional image system, the server has the WebODM \footnote{Drone Mapping Software: \url{www.opendronemap.org/webodm/}}. This is a web-based tool that uses camera images to create (offline) a three-dimensional representation of the environment. The WebODM provides a scaled point cloud of the environment which can be used to measure the area of operation or holes in a building at risk of collapse. This offers the advantage that conclusions can be drawn and a better approach can be planned \cite{DBLP:journals/corr/abs-1709-00587}.

\begin{figure}[!t]
\centering
\includegraphics[width=0.49\textwidth]{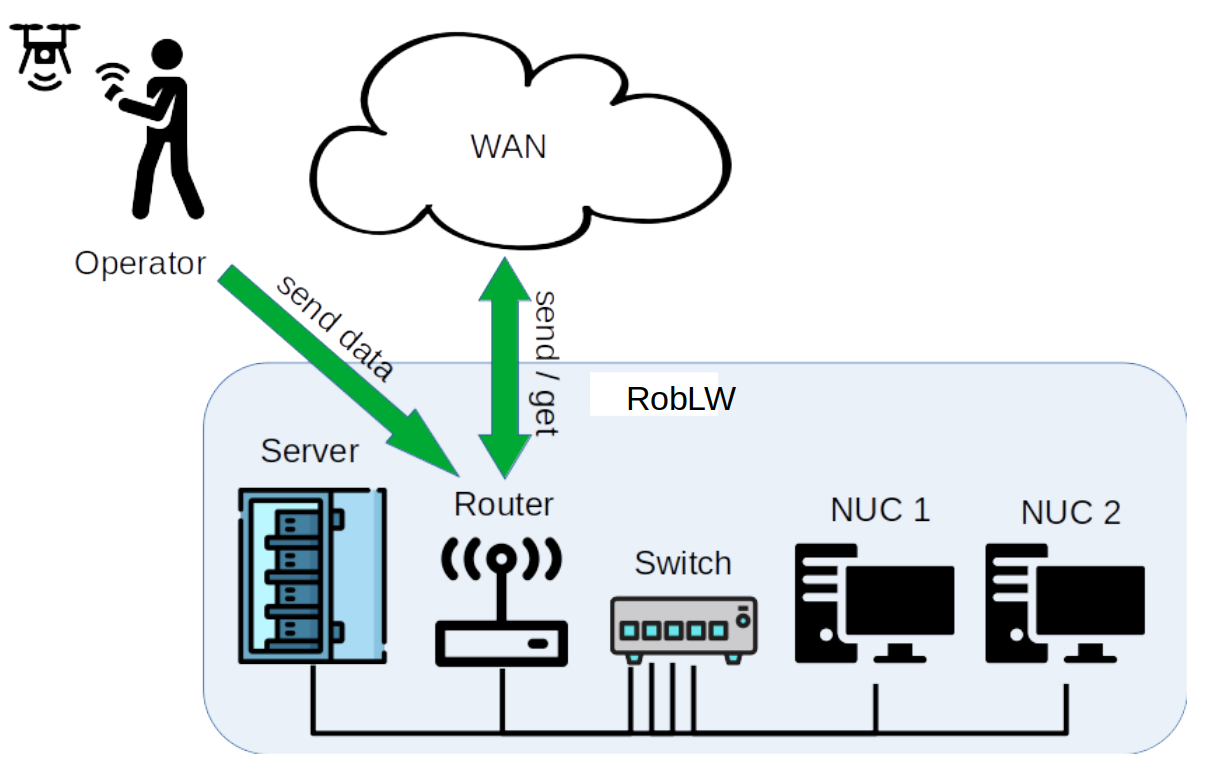}
\caption{RobLW infrastructure. The operator sends the information to RobLW, where it is then processed on the RobLW server. Afterwards, the processed data can be viewed in RobLW on the NUCs or this information can be sent to the control station.}
\label{fig:rlwnetwork}
\end{figure}

\subsubsection{UAVs}
Due to the holes in the hall ceiling caused by the fire, it was assumed that it would be possible to fly UAVs into the building to get an overview of the degree of destruction. A variety of UAVs were brought in for the mission. The DJI Phantom 4 (P4P) was considered due to its maximum flight speed of 72 km/h and stable flying in less than optimal weather. This UAV makes it possible to fly over the entire area in a very short time to get an overview of the current situation on site. However, since the P4P only has a maximum camera resolution of 4000×3000 and a maximum video bit rate of 60 Mbps, the DJI Mavic Pro 2 (MP2) was taken along for detailed recordings. This UAV has a camera resolution of 5472×3648 and a maximum video bit rate of 100 Mbps. Both UAVs can capture 360° panorama images autonomously.

The two UAVs are relative large with a diagonal of 60cm, which is why the DJI Spark and the DJI Mavic Mini were taken along for narrower indoor spaces. These small UAVs were to be used when the destroyed hall had many cables or other broken objects hanging from the ceiling, making it impossible to fly safely in the hall with the MP2 and the P4P. The Spark can take panorama images but not the Mavic Mini.

\subsubsection{Cameras}
Panoramas provide the most information in an image, so mounts have been developed for the P4P and MP2. With these mounts it is possible to attach 360° cameras to the UAVs. The cameras that were available are the Insta 360 One X and the Insta 360 Evo. These have a video function that provides maximum information of the environments through minimal movements of the UAVs. Both cameras have a video resolution of 5760x2880 (15.8MP) at 30 fps.

\subsection{Situation overview model}

\begin{figure}[!t]
\centering
\includegraphics[width=0.49\textwidth]{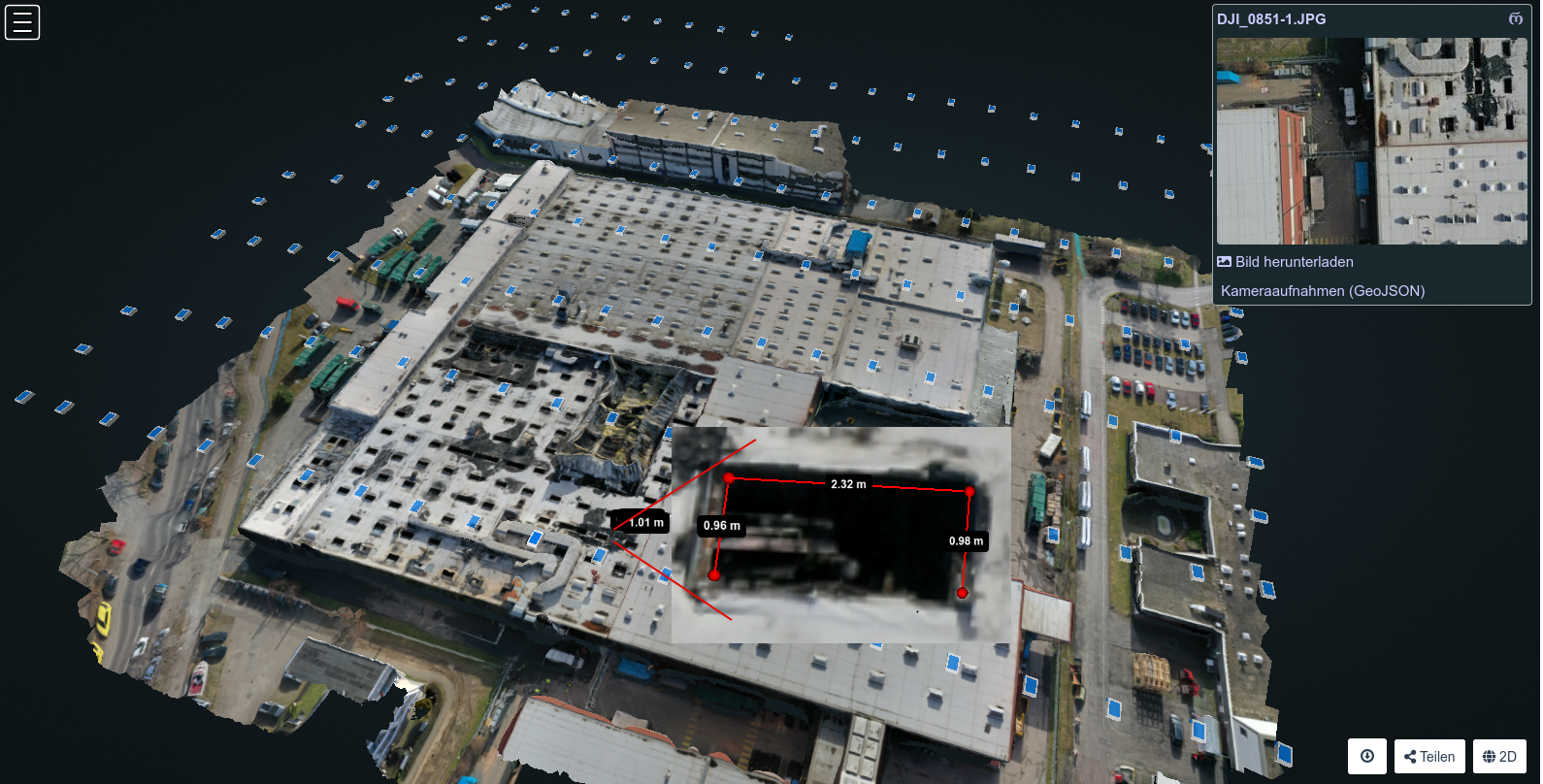}
\caption{Result of the first action is a georeferenced 3D point cloud. It is calculated out of 167 images acquired in a 12 minutes flight of the Mavic at a height of 45 meters. The calculation of the model in the RobLW needs 15 minutes. The measurements taken in the model are used to plan the second action.}
\label{fig:firstaction}
\end{figure}

Since the hall represents unknown terrain, an overview was needed at the beginning. For this purpose, a meander flight with the MP2 was made in a height of 45 meter over the railing in which the hall is located. The UAV flies over the area and creates orthogonal images from the air at regular intervals. The spacing between the images can be adjusted by setting the overlap of the images. In this deployment, an overlap of 70\% was set, generating 167 images from the UAV. These images were then processed using the WebODM tool. WebODM uses Structure from Motion (SfM) and Multi View Stereo (MVS) to create a scaled 3D point cloud that can be viewed in a web browser. Figure \ref{fig:firstaction} shows the point cloud as well as the camera positions, in blue rectangles, over the terrain. By selecting a blue rectangles the image can be displayed in its original resolution (top right). Through the scaled point cloud, the openings of the burned roof hatches could be measured, which have a width of 1m and a length of 2.3m.

\subsection{First flight in - Connection problems with the drone}

\begin{figure*}
\centering
\includegraphics[width=16.8cm,height=23.8cm,keepaspectratio]{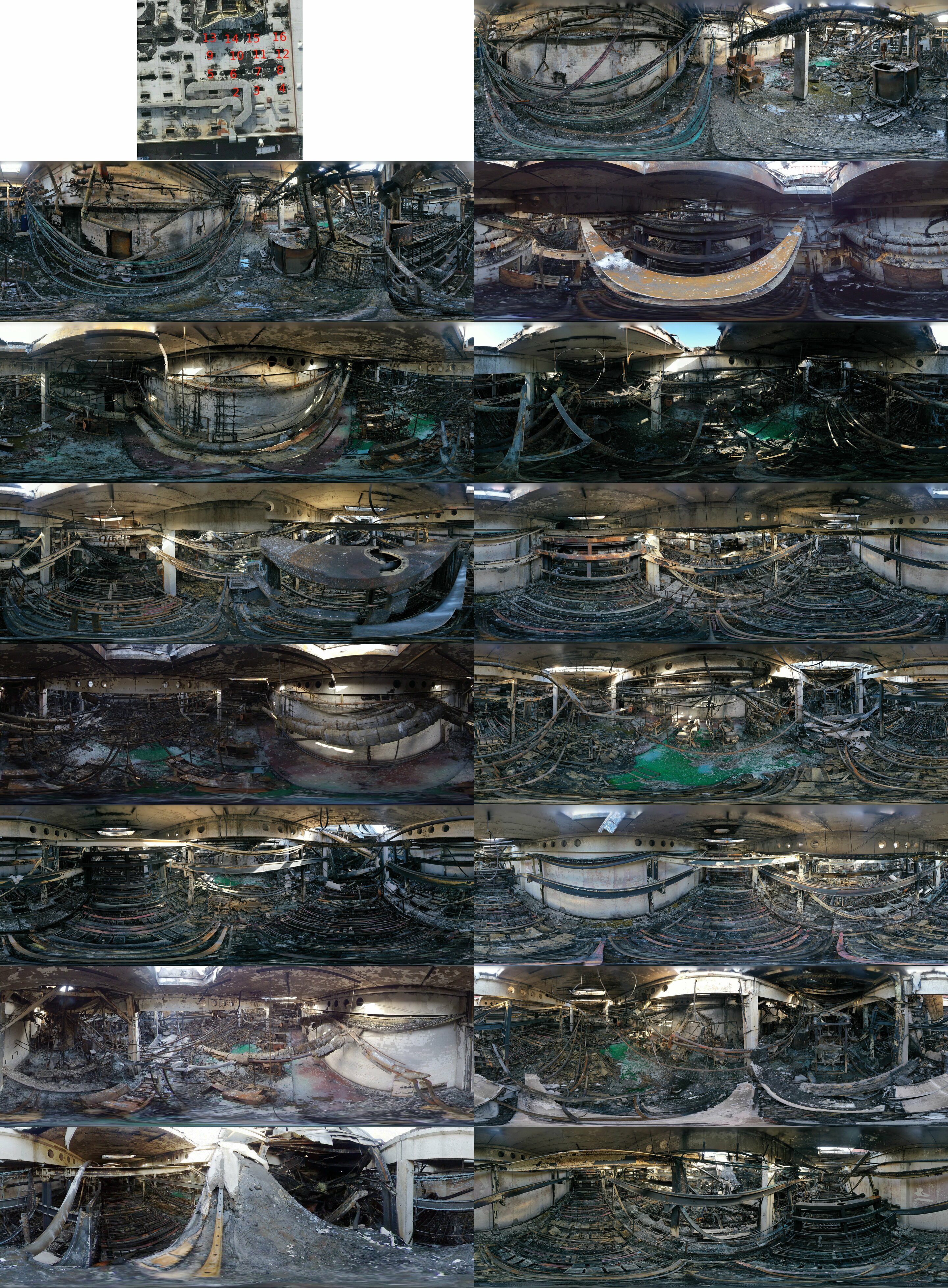}
\caption{360 degree panorama images of the 15 positions (2 - 16).}
\label{fig:pano}
\end{figure*}

From the openings measured above, it was obvious that flying through the ceiling opening into the hall was possible with the P4P and the MP2. There are two problems with this. The small aperture angle of the UAV camera (FOV about 80°) does not allow to see the boundaries of the opening while flying through it. The second problem is to fly the UAV back out of the hall, because the UAV cameras can be pointed forward and downward but not upward. To ensure this, we took the "two pilots approach". While the first UAV enters the hall, the second UAV is positioned exactly above the entry point and assists in flying in. While the first UAV autonomously creates the panoramas in the hall, the second UAV waits above the opening, and after completing the shots, the second pilot navigates the first UAV back out through the opening. For the fly-in, the MP2 was chosen due to the slightly better camera resolution and was equipped with propeller guards. Upon approaching the openings, it was noticeable that the metal of the roof hatches had been severely bent by the fire. This significantly reduced the width of the holes and made it very difficult to fly into the hall. Some openings, for example numbers 4 and 13, could not be flown at all (Fig. \ref{fig:pano}).
To test the "two pilots approach", a particularly large opening was first selected and the two UAV pilots were positioned next to the RobLW, right next to the hall. However, this meant that the radio signal from the remote control had to pass through several layers of concrete and steel, causing the radio link to the MP2 to constantly break down. Normally, if the radio link was lost, the UAV would activate the "Return to Home" (RTH) function, which would cause the UAV to fly back to the launch point on its own. However, the loss of radio link was anticipated, so the safety function was disabled. Furthermore, it is difficult to generate targeted photos without a radio connection. The UAV was able to fly in and out again, but the attempt was aborted due to the difficulties.

\subsection{Second flight in - More stable connection to the drone}

\begin{figure}[!t]
\centering
\includegraphics[width=0.49\textwidth]{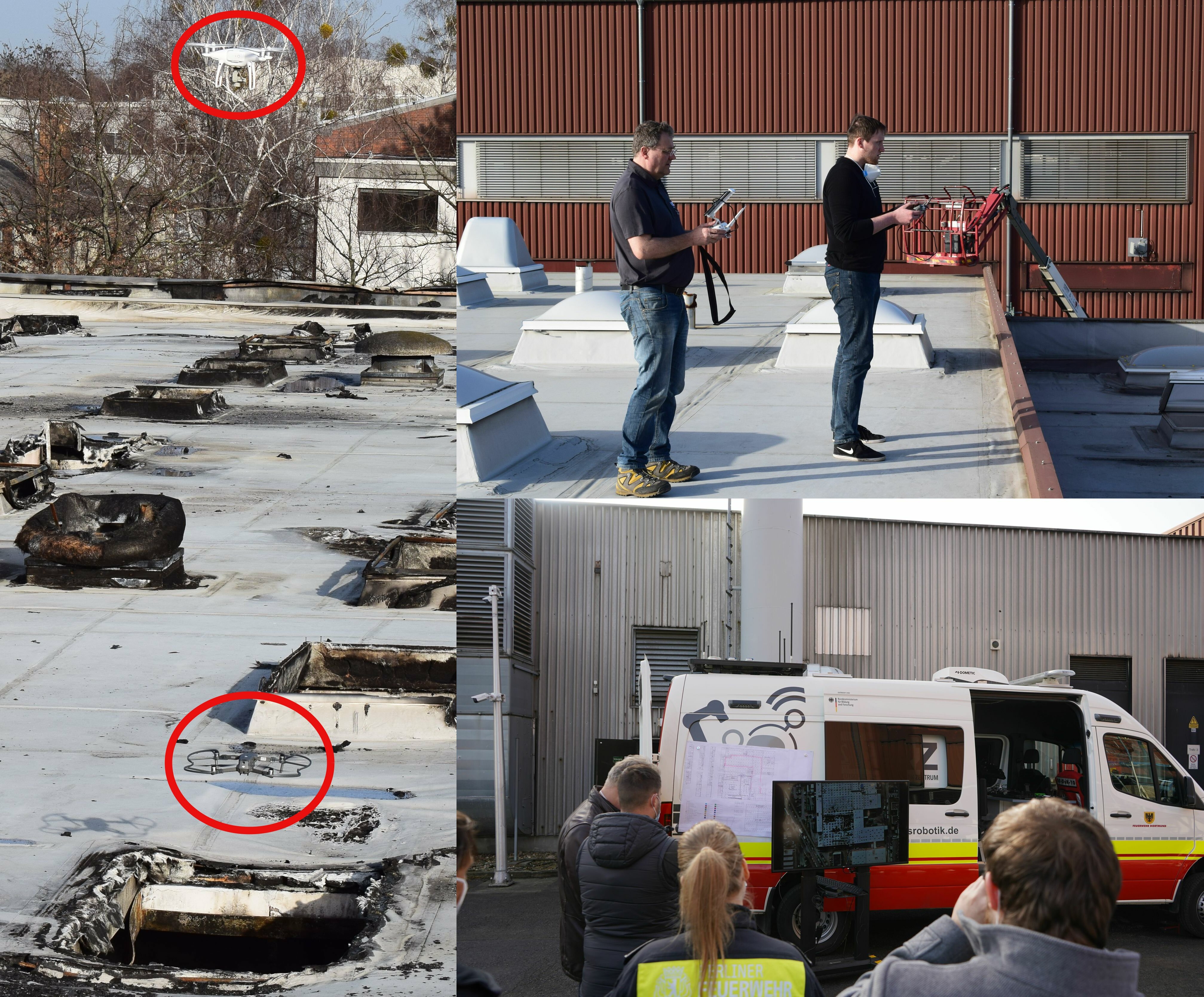}
\caption{Positioning of the two UAVs. The lower one penetrates the damage structure and takes 26 images. The UAVs are controlled by a team of operators and live data is transmitted to the RobLW.}
\label{fig:twouav}
\end{figure}

The frequent disconnections caused an increased risk of crashing the MP2. Therefore, efforts were made to minimize, if not eliminate, the disconnects. The burned hall bordered on an neighboring hall that was accessible. It was located higher than the burned roof which in addition to the visual support from the P4P allowed the pilot to better assess the area. Figure \ref{fig:twouav} left shows the manoeuvre of the two UAVs. During this flight manoeuvre it was noticeable that the lower UAV, the MP2, became more unstable the closer the P4P came. This could be attributed to the P4P's propellers swirling the air too much and causing turbulence for the MP2. In addition Figure \ref{fig:twouav} shows the two UAV pilots in the upper right portion of the image and the live stream shown at RobLW in the lower right portion of the image. The elevated position meant that the radio signal no longer had to pass through several layers of concrete and steel, which meant that the radio signal did not break off when flying into the hall. This meant that the radio signal could be locked in at any time, allowing panoramas to be created more quickly in the hall. 
When panorama function is executed, the UAV creates autonomously 26 individual images (12 MP) of the surroundings, which are then converted to a 360-degree panorama with a resolution of 32 MP (on board). Creating a panorama takes about 1 minute. By post-processing the 26 images at the RobLW, panoramas with 128 MP could be generated. For safety reasons, the MP2 was flown only 1.5m into the hall.

\subsection{Use of panoramic camera - Gathering of missing data}

\begin{figure}[!t]
\centering
\includegraphics[width=0.49\textwidth]{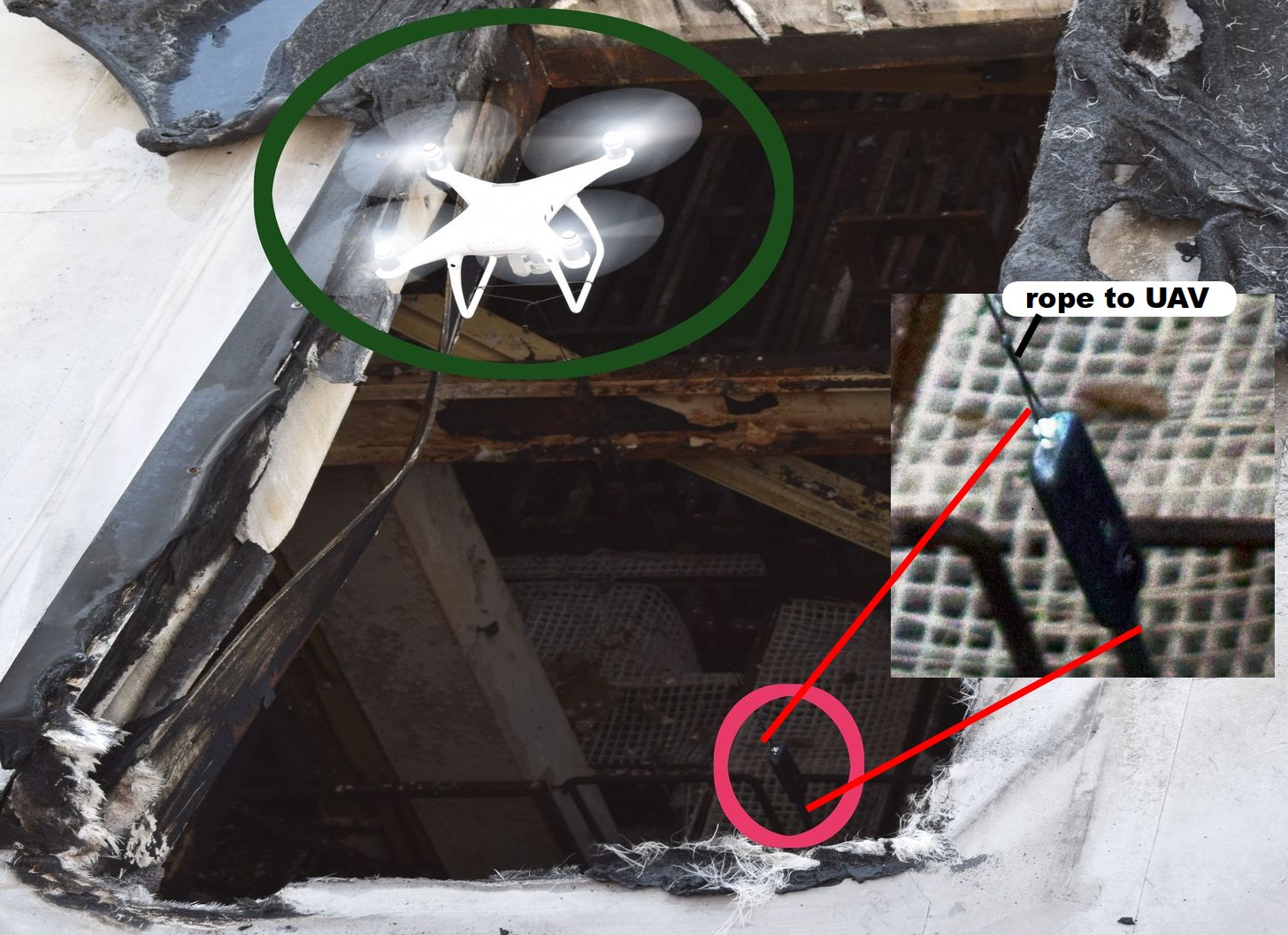}
\caption{360° camera connected to the P4P with a thin rope while inspecting the hall.}
\label{fig:uavrope}
\end{figure}

\begin{figure}[!t]
\centering
\includegraphics[width=0.49\textwidth]{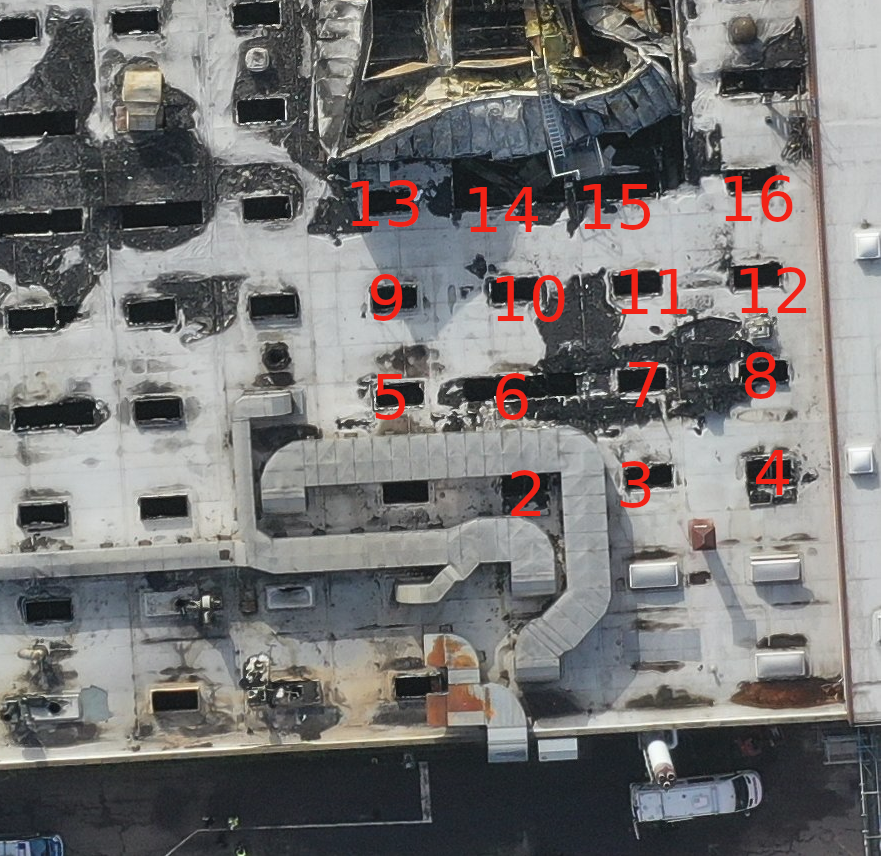}
\caption{Overview image after the initial meander mapping flight. The numbers indicate position of penetration points for the UAV (skylight windows) and therefore insight panorama 360 degree images. Bottom right is the position of the RobLW. The position of the operators were first at the RobLW and second at the neighbouring roof starting directly at the right sight of the image. The line at the right side is a wall which separates the halls. The part right from the line was not effected by the fire.}
\label{fig:overview}
\end{figure}

Another trick was used to capture the positions in the hall that could not be flown directly because the openings were too small. A panoramic camera, namely the Insta 360 One x (110g, 15.8 MP), was attached to the P4P with a 1.5 m long thin rope and inserted into the respective opening from above (Figure \ref{fig:uavrope}). With the Insta 360 One X, it was possible to record 20 minutes of video footage to view all parts of the site. Figure \ref{fig:overview} shows the approached roof hatches into which the camera was lowered with the P4P. The problem with this approach is that the video contains a lot of motion blur due to the moving rope. Furthermore, there are some overexposed and underexposed frames especially when entering the hall from the outside and back. The video has about 36000 frames generated at 30 fps. Furthermore, consecutive frames often show the same or very similar scenes. Therefore, relevant (good) images had to be extracted from the video. The quality of the images is significantly worse than the panoramas taken directly with the UAV (15.8 MP vs. 128 Mp). Related to a field of view of 90° this is 2 MP vs. 16 MP.

\section{Results}

As mentioned above, the video must be post-processed to make important information visible.

\subsection{Dataset filtering by visual SLAM}
To reduce the dataset to the most necessary images, the SLAM framework OpenVSLAM \cite{Sumikura_2019} was used. This framework uses the ORB2-SLAM \cite{Mur_Artal_2017} algorithm and supports equirectangular camera types. With OpenVSLAM each keyframe of the video was saved. This ensures that these images contain representative information of the environment. This method reduced the dataset from 36000 frames to a dataset of over 700 key frames. Furthermore a flight trajectory could be generated and also the high quality panoramas could be localized.

\subsection{Dataset filtering by blur detection}
\begin{figure}[!t]
\centering
\includegraphics[width=0.49\textwidth]{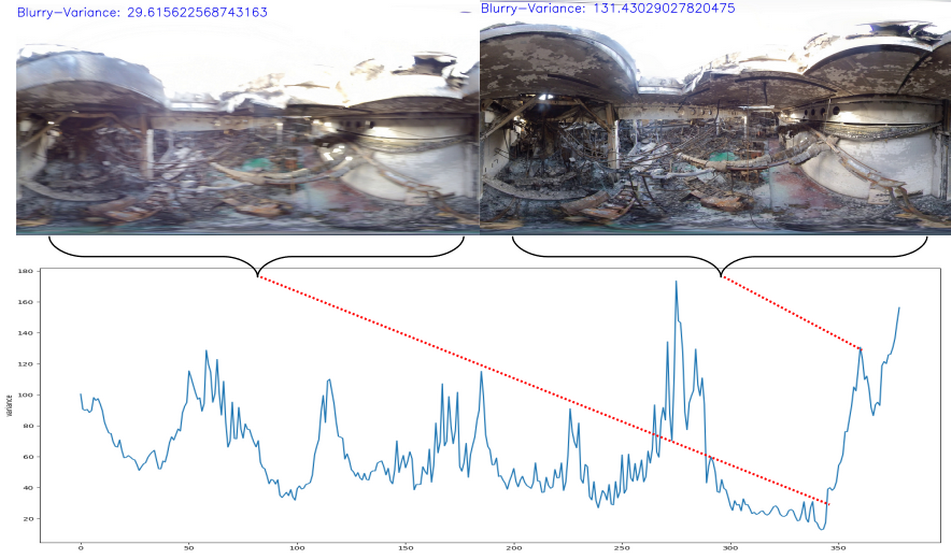}
\caption{Example of blur detection. At the top right a sharp image with a high variance and at the top left an blurred image with a lower variance. Below the corresponding variance curve of the blur detection.}
\label{fig:blurdetection}
\end{figure}

During the key frame calculation of the video data set by visual SLAM, only non-representative images and duplicate images are removed from the original data set. Images with motion blur were not completely removed. Therefore, a Laplace Blur Detection \cite{7894491, cesit21, SIEBERTH20161, 6959928} was used to remove blurred images from the video as well. In this detection, the Laplace filter is applied to the image and then the variance is determined. If the variance is below a certain threshold, then the image should normally be discarded. Figure \ref{fig:blurdetection} serves as an example. However, since the variance changes with the environment and thus the threshold must be adjusted, a dynamic threshold was added to the detector from \cite{7894491}. This takes into account the images before and after the current image \( B_n \). Therefore the variances of \( k \) images and the variance of the current image are calculated. The mean value of these corresponds to the threshold. We have chosen the value 20 for the parameter \( k \). A problem with the procedure is that at the beginning of the algorithm or at the first image no previous images exist, the same problem exists at the end of the data set. Therefore, the parameter \( j \) with the value 0 was introduced, which ensures that non-existing images are not included in the calculation.

\begin{align}
\frac{1}{(k*2) + 1 + j} \sum_{i=n-k}^{n+k} Var(B_i) \text{   } | \text{   } j = \begin{cases}    j+1 & \text{if  } i < 0 \\    j+1 & \text{if  } i >\text{end}\end{cases}
\end{align}

\subsection{3D Viewer}
\begin{figure}[!t]
\centering
\includegraphics[width=0.49\textwidth]{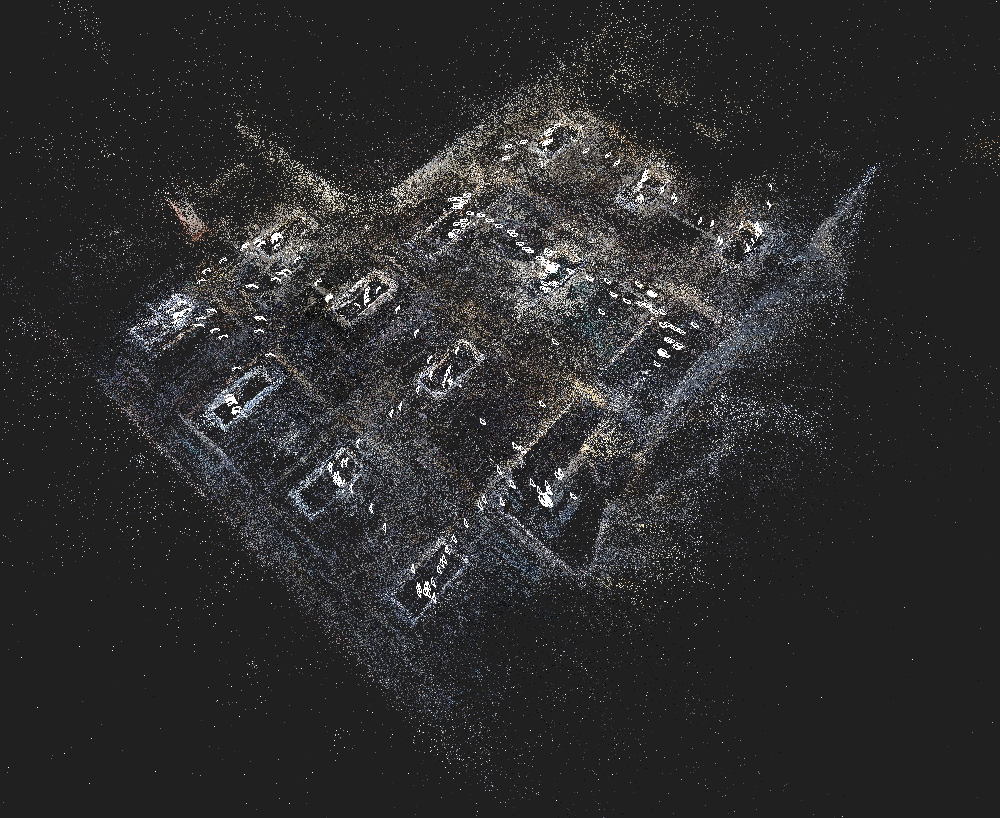}
\caption{Semi dense point cloud with localized image position calculated with structure from motion or visual SLAM.}
\label{fig:semidense}
\end{figure}

\begin{figure}[!t]
\centering
\includegraphics[width=0.49\textwidth]{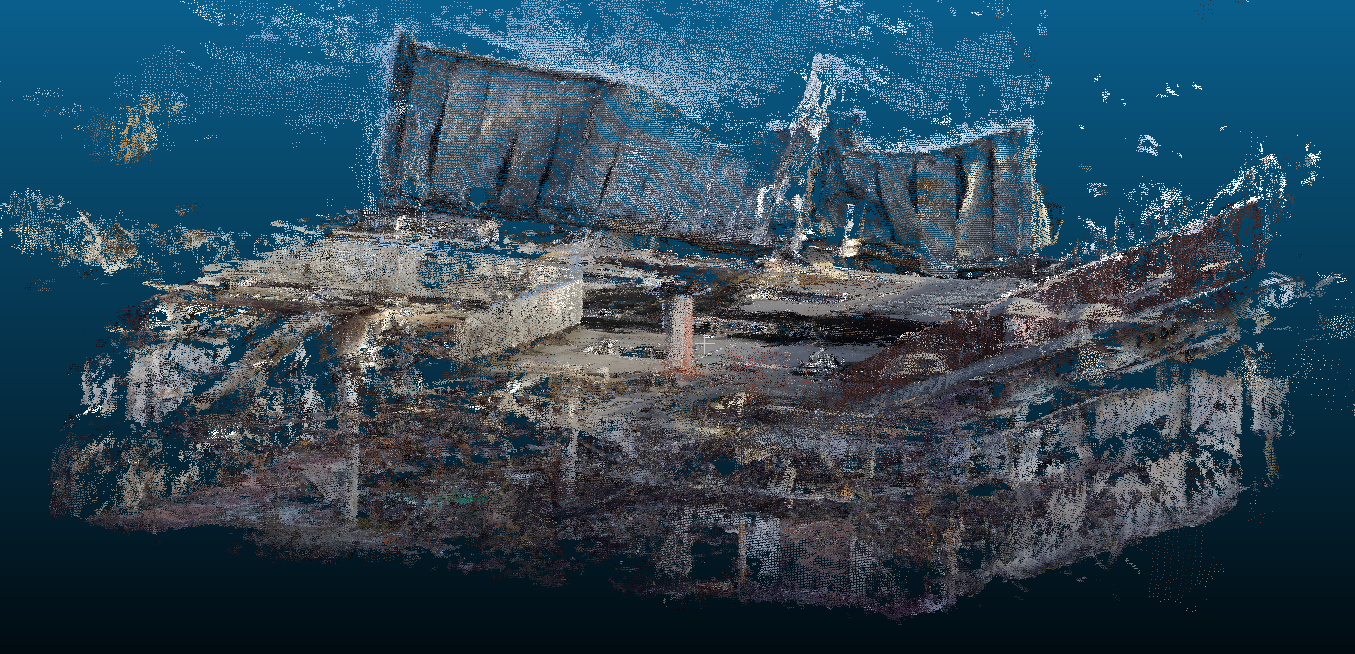}
\caption{Dense point cloud of the structure calculated with multi view stereo.}
\label{fig:dense}
\end{figure}

With the blur detection the dataset was reduced to a little more than 200 images and processed with WebODM. WebODM uses OpenSfM \cite{ozyesil2017survey} to create semi dense point clouds, as shown in Figure \ref{fig:semidense}. This gives an abstract overview of the environment and also localizes the camera positions, which can be used in a 2D viewer. 
The dense point cloud was created using OpenMVS \cite{CGV-052} and provides detailed representation of the environment (figure \ref{fig:dense}). This detailed representation could be used as an overview to get an overall impression of the situation.

\subsection{360° panorama viewer}
\begin{figure}[!t]
\centering
\includegraphics[width=0.49\textwidth]{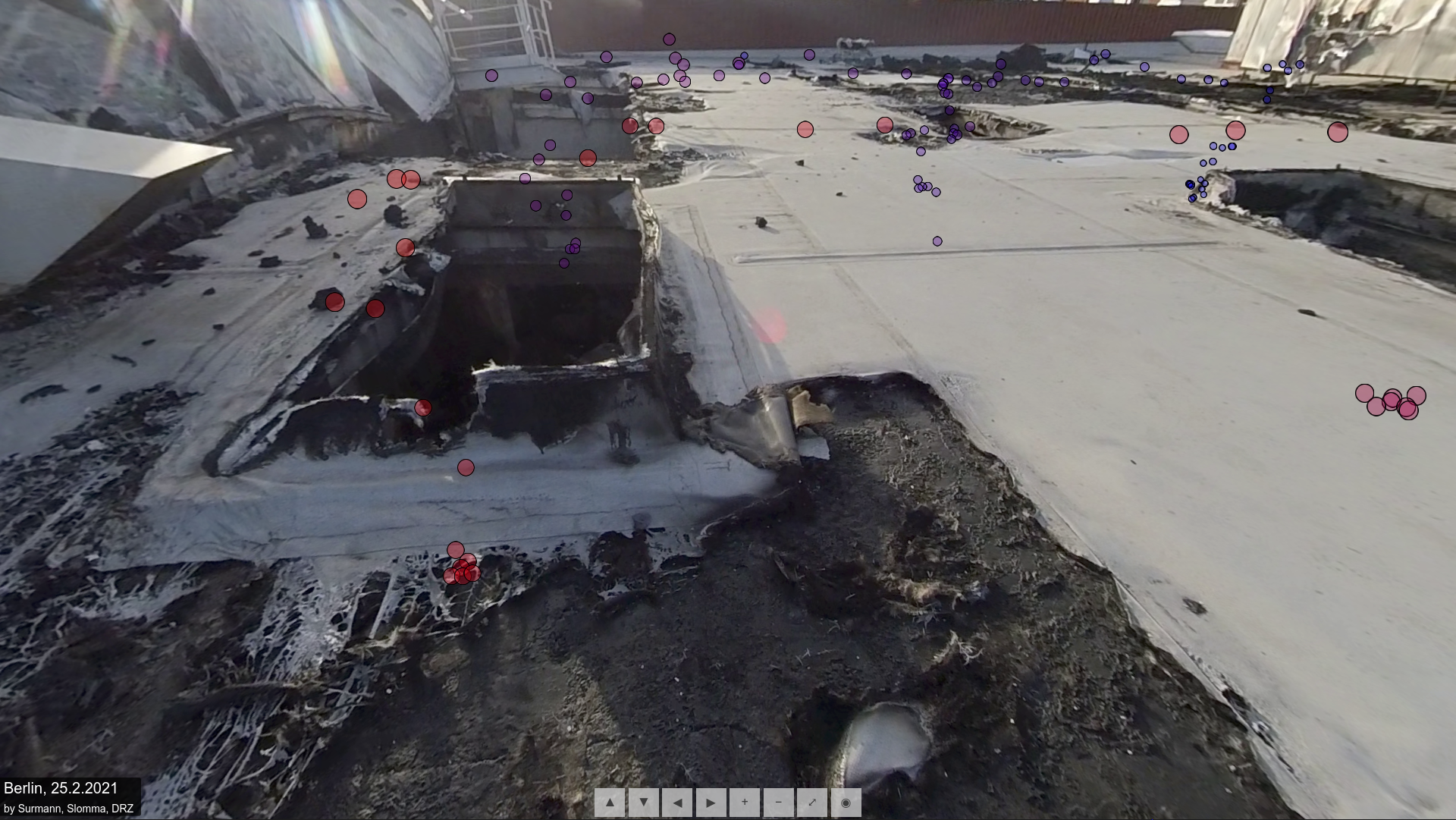}
\caption{Panorama viewer. The user can rotate, zoom and navigate in the images. The red to blue circles indicate further 3D panorama images. Large and red mean near by images, small and blue mean far away images according to the localization in the semi dense point cloud.}
\label{fig:panoviewer}
\end{figure}

Based on the localization from OpenSfM, an interactive panorama viewer was implemented. It uses the Pannelum web viewer javascript plugin \footnote{\url{pannellum.org/}}. With this it is possible to view 360° panoramas interactively. The viewer was extended by the localization information (reconstruction.json) from OpenSfM, so that it is possible to switch between the individual panoramas. As shown in Figure \ref{fig:panoviewer}, there are several color-coded circles. These represent the localized panoramas. The more red a point is, the closer the panorama is to the current position. If the panorama is further away, it will be blue. By clicking into the circle the next panorama appears. The presentation was inspired by google street view.

\section{CONCLUSIONS}
This article reports on the use of the UAV in Berlin in February 2021 to inspect an industrial hall with hazardous substances that was in danger of collapsing after a large fire. It has been shown that the use of the panorama function of the UAV provides optimal information of the environment, but there is always a great risk that the UAV collides with the environment. Therefore, the use of a 360 degree camera, which is let into the hall, is also suitable. This allows image information to be collected from different heights without the risk of crashing. The disadvantage of the camera is the post-processing of the data. These have to be freed from non-optimal image information in order to create a suitable 3D model of the environment. It also turned out that 3D models are interesting for the robots, but not for the emergency services. They have difficulties to find their way in a 3D model and to extract the necessary information from it. Therefore, the 3D model should only serve as a rough overview and the implemented panorama viewer was used by the emergency forces to obtain the necessary information from the environment. Finally, thanks to the unknown reviewers whose questions we will try to answer below.

\begin{enumerate}
    \item {\bf After the deployment, how about the decontamination of the UAVs and Insta 360?} 
    After the mission, the UAVs and the 360° cameras were carefully cleaned. As expected, a final measurement showed no decontamination, as both the UAVs and the camera had no direct contact with the environment. 

\item {\bf What information the panorama viewer can obtain and the 3D model cannot?}
Due to the high resolution of the single panorama images they contain a lot more details compared to the dense point clouds. The calculation of dense point clouds by multi view stereo algorithm downscale the image i.e. by a number of 4. Furthermore, has the used PatchMatch Algorithm difficulties with non textured surfaces. Through the panorama viewer it was possible to view individual panoramas and through their high resolution it was possible to zoom into interesting areas to get a detailed view of the individual areas as well as directly move to other places. 

\item {\bf Which types of analyzed data would be finally useful for polices further investigations and when was the timing of its provision?}
Both the point clouds as well as the panorama images are useful and important. The dense point clouds give a spatial impression of the operation and serve for the localization of the panoramas as well as for the measurements. The panoramas in turn contain the details for assessing the statics and geometry of the building as well as the cause of the fire. The outdoor 3D overview model was calculated at the side (in the RobLW) and provided 30 Minutes after the UAV flight. It served for the planning of the further proceeding. The panoramas were also shown directly after the flights during the mission. The 3D interior model and the final browser-based panoramic viewer were delivered online 3 days after the mission.  

\item {\bf How the results helped for the further investigation?}
Due to the ban on entering the hall after the fire, there was no information from the inside of the hall, neither regarding the statics and danger of collapse, nor regarding the cause of the fire. The images provided are the basis for pumping out the decontaminated firefighting water, assessing the statics and planning the deconstruction as well as determining the cause of the fire.

\item {\bf Before the deployment, the UAV team had training for a similar application? If so, what's new in the real deployment comparing to the training?}
The UAV team had already trained flying inside and outside buildings and flying into buildings through windows. In addition, there was already operational experience with flying robots during earthquakes, e.g. (Amatrice / Italy 2016). Furthermore, there was extensive experience in the development and application of algorithms from the field of image processing and computer science. Flying in through skylights instead of windows was new. Furthermore, the images are new and different from those of earthquakes. Fire, especially in industrial halls leads to massive destruction of the environment. 

\item {\bf Are there any improvements in the deployment of these drones?}
Not directly in the use of UAVs, but in the development of the image processing and display algorithms. This refers on the one hand to the better display (PanoViewer) and on the other hand to the processing speed for data evaluation. 

\end{enumerate}

\begin{figure}[!t]
\centering
\includegraphics[width=0.49\textwidth]{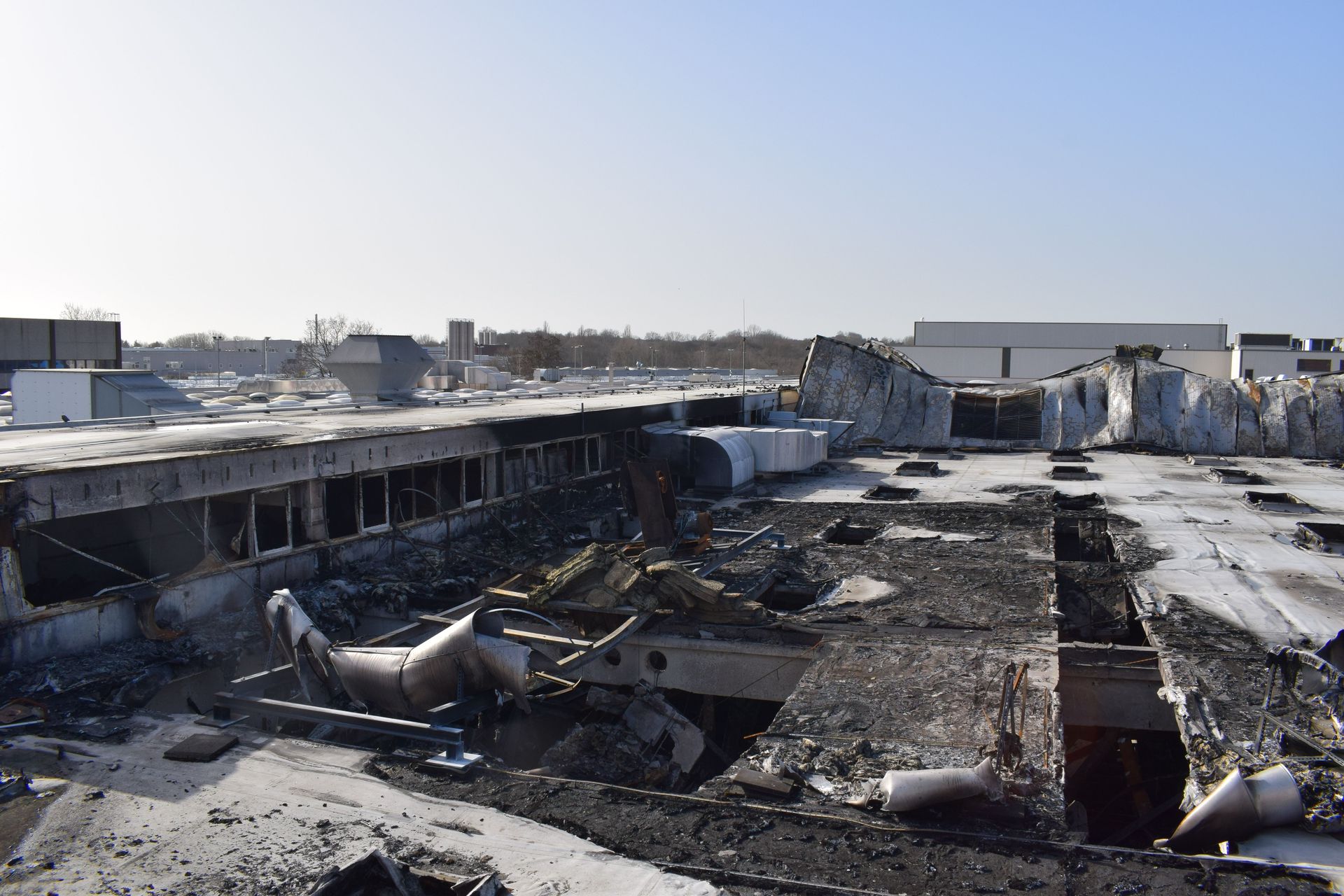}
\includegraphics[width=0.49\textwidth]{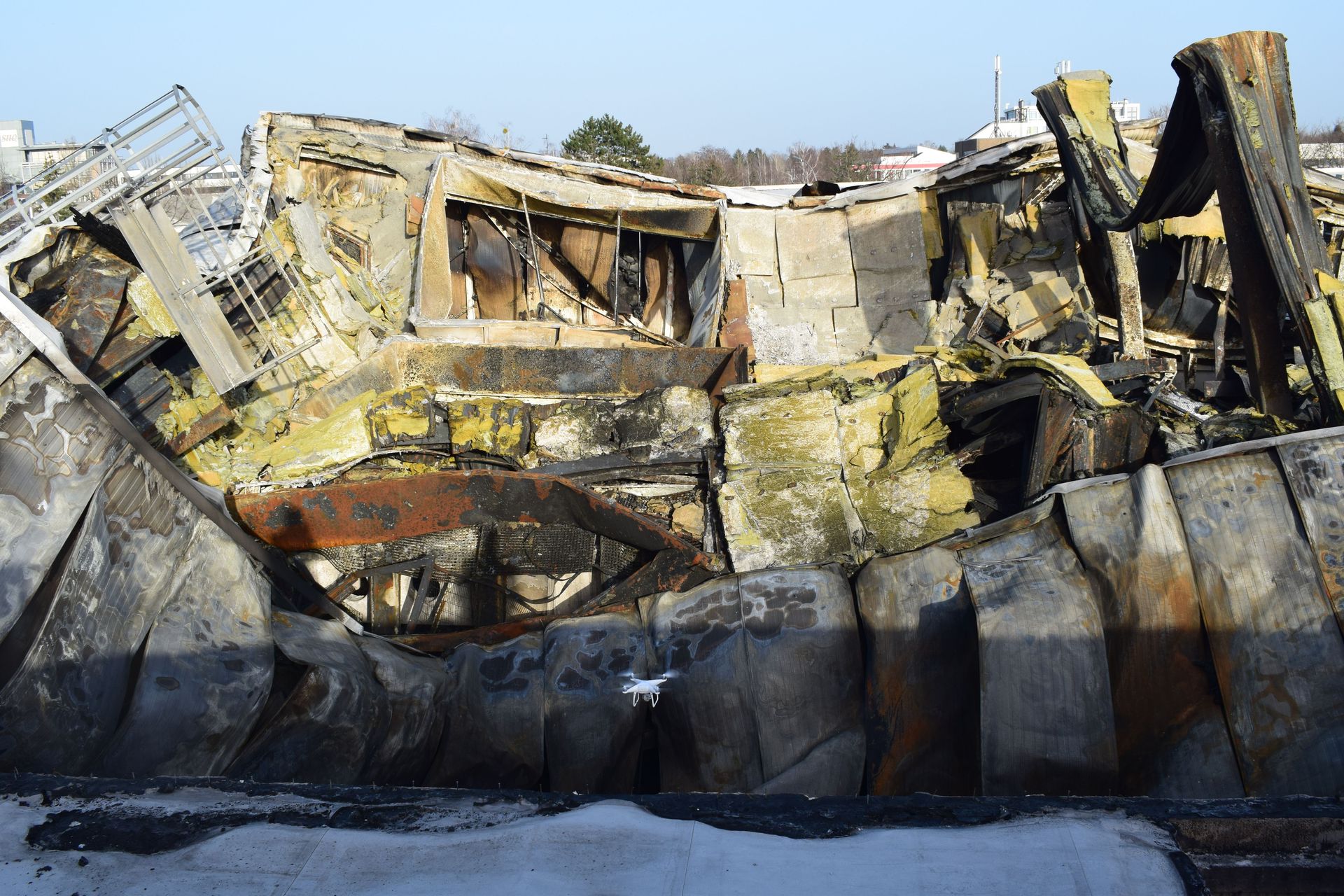}
\caption{Detail view of the roof of the industrial hall}
\label{fig:details}
\end{figure}

\addtolength{\textheight}{-12cm}   

 


\bibliographystyle{IEEEtran} 
\bibliography{IEEEabrv,literatur} 

\begin{thebibliography}{10}
\providecommand{\url}[1]{#1}
\csname url@rmstyle\endcsname
\providecommand{\newblock}{\relax}
\providecommand{\bibinfo}[2]{#2}
\providecommand\BIBentrySTDinterwordspacing{\spaceskip=0pt\relax}
\providecommand\BIBentryALTinterwordstretchfactor{4}
\providecommand\BIBentryALTinterwordspacing{\spaceskip=\fontdimen2\font plus
\BIBentryALTinterwordstretchfactor\fontdimen3\font minus
  \fontdimen4\font\relax}
\providecommand\BIBforeignlanguage[2]{{%
\expandafter\ifx\csname l@#1\endcsname\relax
\typeout{** WARNING: IEEEtran.bst: No hyphenation pattern has been}%
\typeout{** loaded for the language `#1'. Using the pattern for}%
\typeout{** the default language instead.}%
\else
\language=\csname l@#1\endcsname
\fi
#2}}

\bibitem{berlinfeuerwehr}
\BIBentryALTinterwordspacing
B.~Feuerwehr, ``Großbrand in {S}törfallbetrieb in {B}erlin-{M}arienfelde,''
  Berlin, Germany, 11.02.2021. [Online]. Available:
  \url{https://www.berliner-feuerwehr.de/aktuelles/einsaetze/grossbrand-in-stoerfallbetrieb-in-berlin-marienfelde-3719}
\BIBentrySTDinterwordspacing

\bibitem{5981550}
T.~Linder, V.~Tretyakov, S.~Blumenthal, P.~Molitor, D.~Holz, R.~Murphy,
  S.~Tadokoro, and H.~Surmann, ``Rescue robots at the collapse of the municipal
  archive of cologne city: A field report,'' in \emph{2010 IEEE Safety Security
  and Rescue Robotics}, July 2010, pp. 1--6.

\bibitem{japan}
F.~Matsuno, N.~Sato, K.~Kon, H.~Igarashi, T.~Kimura, and R.~Murphy,
  ``Utilization of robot systems in disaster sites of the great eastern japan
  earthquake,'' vol.~92, pp. 1--17, 12 2014.

\bibitem{Kruijff-amatrice}
I.~{Kruijff-Korbayová}, L.~{Freda}, M.~{Gianni}, V.~{Ntouskos}, V.~{Hlaváč},
  V.~{Kubelka}, E.~{Zimmermann}, H.~{Surmann}, K.~{Dulic}, W.~{Rottner}, and
  E.~{Gissi}, ``Deployment of ground and aerial robots in earthquake-struck
  amatrice in italy (brief report),'' in \emph{2016 IEEE International
  Symposium on Safety, Security, and Rescue Robotics (SSRR)}, Oct 2016, pp.
  278--279.

\bibitem{6523866}
G.-J.~M. Kruijff, F.~Pirri, M.~Gianni, P.~Papadakis, M.~Pizzoli, A.~Sinha,
  V.~Tretyakov, T.~Linder, E.~Pianese, S.~Corrao, F.~Priori, S.~Febrini, and
  S.~Angeletti, ``Rescue robots at earthquake-hit mirandola, italy: A field
  report,'' in \emph{2012 IEEE International Symposium on Safety, Security, and
  Rescue Robotics (SSRR)}, Nov 2012, pp. 1--8.

\bibitem{advanced-robotics-2014}
G.~Kruijff, I.~Kruijff-Korbayov{\'a}, S.~Keshavdas, B.~Larochelle,
  M.~Jan{\'i}{\v c}ek, F.~Colas, M.~Liu, F.~Pomerleau, R.~Siegwart,
  M.~Neerincx, R.~Looije, N.~Smets, T.~Mioch, J.~van Diggelen, F.~Pirri,
  M.~Gianni, F.~Ferri, M.~Menna, R.~Worst, T.~Linder, V.~Tretyakov, H.~Surmann,
  T.~Svoboda, M.~Rein{\v s}tein, K.~Zimmermann, T.~Pet{\v r}{\'i}{\v c}ek, and
  V.~Hlav{\'a}{\v c}, ``Designing, developing, and deploying systems to support
  human-robot teams in disaster response,'' \emph{Advanced Robotics}, vol.~28,
  no.~23, pp. 1547--1570, 2014.

\bibitem{7017681}
G.~Steinbauer, J.~Maurer, and A.~Ciossek, ``Field report: Civil protection
  exercise gas storage,'' in \emph{2014 IEEE International Symposium on Safety,
  Security, and Rescue Robotics (2014)}, Oct 2014, pp. 1--2.

\bibitem{9419563}
N.~Sato, ``Evaluation exercise in the seminar for standard test method of
  ground robot at fukushima robot test field,'' in \emph{2021 IEEE
  International Conference on Intelligence and Safety for Robotics (ISR)},
  March 2021, pp. 127--130.

\bibitem{mayer:hal-02128385}
\BIBentryALTinterwordspacing
S.~Mayer, L.~Lischke, and P.~W. Wo{\'z}niak, ``{Drones for Search and
  Rescue},'' in \emph{{1st International Workshop on Human-Drone
  Interaction}}.\hskip 1em plus 0.5em minus 0.4em\relax Glasgow, United
  Kingdom: {Ecole Nationale de l'Aviation Civile [ENAC]}, May 2019. [Online].
  Available: \url{https://hal.archives-ouvertes.fr/hal-02128385}
\BIBentrySTDinterwordspacing

\bibitem{surmann2019ssrr}
H.~Surmann, R.~Worst, T.~Buschmann, A.~Leinweber, A.~Schmitz, G.~Senkowski, and
  N.~Goddemeiner, ``Integration of uavs in urban search and rescue missions,''
  in \emph{2019 IEEE International Symposium on Safety, Security, and Rescue
  Robotics (SSRR)}.\hskip 1em plus 0.5em minus 0.4em\relax IEEE, 2019, pp.
  203--209.

\bibitem{DBLP:journals/corr/abs-1709-00587}
\BIBentryALTinterwordspacing
A.~Gawel, R.~Dub{\'{e}}, H.~Surmann, J.~I. Nieto, R.~Siegwart, and C.~Cadena,
  ``3d registration of aerial and ground robots for disaster response: An
  evaluation of features, descriptors, and transformation estimation,''
  \emph{CoRR}, vol. abs/1709.00587, 2017. [Online]. Available:
  \url{http://arxiv.org/abs/1709.00587}
\BIBentrySTDinterwordspacing

\bibitem{Sumikura_2019}
\BIBentryALTinterwordspacing
S.~Sumikura, M.~Shibuya, and K.~Sakurada, ``Openvslam,'' \emph{Proceedings of
  the 27th ACM International Conference on Multimedia}, Oct 2019. [Online].
  Available: \url{http://dx.doi.org/10.1145/3343031.3350539}
\BIBentrySTDinterwordspacing

\bibitem{Mur_Artal_2017}
\BIBentryALTinterwordspacing
R.~Mur-Artal and J.~D. Tardos, ``Orb-slam2: An open-source slam system for
  monocular, stereo, and rgb-d cameras,'' \emph{IEEE Transactions on Robotics},
  vol.~33, no.~5, p. 1255–1262, Oct 2017. [Online]. Available:
  \url{http://dx.doi.org/10.1109/TRO.2017.2705103}
\BIBentrySTDinterwordspacing

\bibitem{7894491}
R.~Bansal, G.~Raj, and T.~Choudhury, ``Blur image detection using laplacian
  operator and open-cv,'' in \emph{2016 International Conference System
  Modeling Advancement in Research Trends (SMART)}, Nov 2016, pp. 63--67.

\bibitem{cesit21}
R.~A. Pagaduan., M.~C. {R. Aragon}., and R.~P. Medina., ``iblurdetect: Image
  blur detection techniques assessment and evaluation study,'' in
  \emph{Proceedings of the International Conference on Culture Heritage,
  Education, Sustainable Tourism, and Innovation Technologies - CESIT,},
  INSTICC.\hskip 1em plus 0.5em minus 0.4em\relax SciTePress, 2021, pp.
  286--291.

\bibitem{SIEBERTH20161}
\BIBentryALTinterwordspacing
T.~Sieberth, R.~Wackrow, and J.~H. Chandler, ``Automatic detection of blurred
  images in uav image sets,'' \emph{ISPRS Journal of Photogrammetry and Remote
  Sensing}, vol. 122, pp. 1--16, 2016. [Online]. Available:
  \url{https://www.sciencedirect.com/science/article/pii/S0924271616303999}
\BIBentrySTDinterwordspacing

\bibitem{6959928}
B.~T. Koik and H.~Ibrahim, ``A literature survey on blur detection algorithms
  for digital imaging,'' in \emph{2013 1st International Conference on
  Artificial Intelligence, Modelling and Simulation}, 2013, pp. 272--277.

\bibitem{ozyesil2017survey}
O.~Ozyesil, V.~Voroninski, R.~Basri, and A.~Singer, ``A survey of structure
  from motion,'' 2017.

\bibitem{CGV-052}
\BIBentryALTinterwordspacing
Y.~Furukawa and C.~Hernández, ``Multi-view stereo: A tutorial,''
  \emph{Foundations and Trends® in Computer Graphics and Vision}, vol.~9, no.
  1-2, pp. 1--148, 2015. [Online]. Available:
  \url{http://dx.doi.org/10.1561/0600000052}
\BIBentrySTDinterwordspacing

\end{thebibliography}


\end{document}